\documentclass[letterpaper, 10 pt, conference]{ieeeconf}

\usepackage[T1]{fontenc}    
\usepackage[utf8]{inputenc} 

\usepackage{url}            
\usepackage{booktabs}       
\usepackage{amsfonts}       

\usepackage[dvipsnames,svgnames]{xcolor} 

\usepackage{amsmath} 

\usepackage{amsthm}
\usepackage{amssymb} 
\usepackage{mathtools}
\usepackage{mathrsfs}
\usepackage{mathdots}
\usepackage{bm}

\usepackage{graphics} 
\usepackage{tikz}
\usetikzlibrary{shapes, shapes.geometric, arrows, calc}
\usetikzlibrary{bayesnet}
\usetikzlibrary{fadings}
\usetikzlibrary{patterns}
\usepackage{yhmath}
\usepackage{cancel}
\usepackage{gensymb}

\usepackage{pgfkeys}
\usepackage{standalone}
\usepackage{fontawesome5} 

\usepackage{subcaption}
\usepackage{cite}
\usepackage{graphicx}
\usepackage{multirow}
\usepackage{threeparttable}
\usepackage{adjustbox}
\usepackage{aligned-overset}
\usepackage{dblfloatfix}

\usepackage{float}
\usepackage{verbatim}
\let\labelindent\relax
\usepackage[shortlabels]{enumitem}
\usepackage{array}

\usepackage{hyperref} 
\hypersetup{
 colorlinks = true,
 linkcolor=black,
 urlcolor=Blue}

\definecolor{revise}{rgb}{0.54, 0.17, 0.89} 

\IEEEoverridecommandlockouts                              

\usepackage{algorithm}
\usepackage{amsmath}
\usepackage{algorithmic}
\usepackage{amsfonts}
\usepackage{siunitx}
\usepackage{textcomp}
\usepackage{graphicx}
\usepackage{subcaption}

\title{Fast Planning for Multi-object Multi-target Throwing}
\author{Zhengming Zhu$^{*}$ \quad Yang Liu$^{*}$ \quad Xiao Gao \quad Aude Billard
\thanks{$^{*}$Equal contribution. The authors are with the Learning Algorithms and Systems Laboratory (LASA), EPFL, Switzerland. 
Corresponding author: Yang Liu 
({\tt\footnotesize yangliudh@gmail.com}).}
}

\begin{document}
\maketitle

\begin{abstract}
Robot throwing has emerged as a promising technique for improving efficiency in logistics and warehouse automation, by enlarging the workspace and speeding up the process. To significantly increase the throwing system's throughput, we develop strategies for throwing multiple objects in one swipe. Such multi-object multi-target throwing (MOMT) leverages the large degrees of freedom of anthropomorphic hands. The key is to quickly generate fast and feasible throwing motions, which involves a complex trade-off between short trajectory duration and short planning time. We solve this problem in two stages. Offline, we build a model of the feasible set by combining object's inverted flying dynamics and the robot's kinematics and dynamics. Online, we generate feasible throws through fast solution matching and filtering of object's valid detach state and robot's feasible state that can compose sequences of throws in less than 5~ms. We validate the framework on a 7-DoF manipulator equipped with a multi-fingered hand. In simulation, coordinated two-object throwing reduces execution time by up to \SI{46}{\percent} compared to independent single-object planning, and this improvement is maintained when scaling to three objects. Real-world experiments with two objects confirm a \SI{29}{\percent} reduction; the remaining gap to the theoretical \SI{50}{\percent} is attributed to inter-throw transition overhead. When target positions are randomly changed mid-execution, the system re-plans and successfully reaches the new targets within 100~ms latency without stopping the robot. These results establish the first unified planning framework for MOMT throwing---demonstrating scalability to multiple objects in simulation and real-world feasibility on two-object tasks---advancing the frontier of high-throughput robotic manipulation. A video summarizing the proposed method and the hardware experiments is available at \href{https://liuyangdh.github.io/momt-video/}{https://liuyangdh.github.io/momt-video}.
\end{abstract}

\setlength{\textfloatsep}{8pt} 
\setlength{\abovecaptionskip}{3pt}
\setlength{\belowcaptionskip}{0pt}
\setlength{\intextsep}{5pt}
\section{Introduction}
Robot throwing is a challenging dynamic manipulation task, whereby the robot agilely throws an object toward a target. This approach replaces traditional pick and place, with high potential to boost efficiency in many common applications of robots in industry. Yet, past solutions focus exclusively on single-object throwing, where the robot picks and throws one object at a time in repetition. 

With the growing availability of low-cost multi-fingered anthropomorphic robot hands, robots now possess the capability of holding multiple objects simultaneously. With multiple objects in hand, it becomes possible to synthesize a single stroke of high-speed motion that throws different objects to their designated targets, thereby drastically increasing object transport efficiency. This is particularly advantageous in high-throughput sorting applications where items must be continuously dispatched to spatially distributed targets, such as parcel routing to multiple destination chutes, recycling separation into distinct material bins, and warehouse order consolidation. In such settings, a single multi-target throwing stroke amortizes the arm's transport motion across several items and destinations at once, yielding throughput gains beyond what repeated single-object pick-and-place or single-target throwing can achieve.

However, compared to single-object throwing, {\em multi-object multi-target} (MOMT) throwing presents a unique challenge: MOMT introduces explosive combinatorial complexity when searching for the best (e.g., time-optimal) throwing motion. This substantially increases planning time, which in turn compromises the throughput of the overall throwing system. Given the demanding and unpredictable MOMT throwing task queries common in industrial settings, it is therefore necessary to strategically balance the time spent on motion planning with the duration of the throwing motion. In this work, we seek a planning framework that provides a favorable trade-off in this dilemma.
\begin{figure}[t!]
    \centering
    \includegraphics[width=1.0\linewidth]{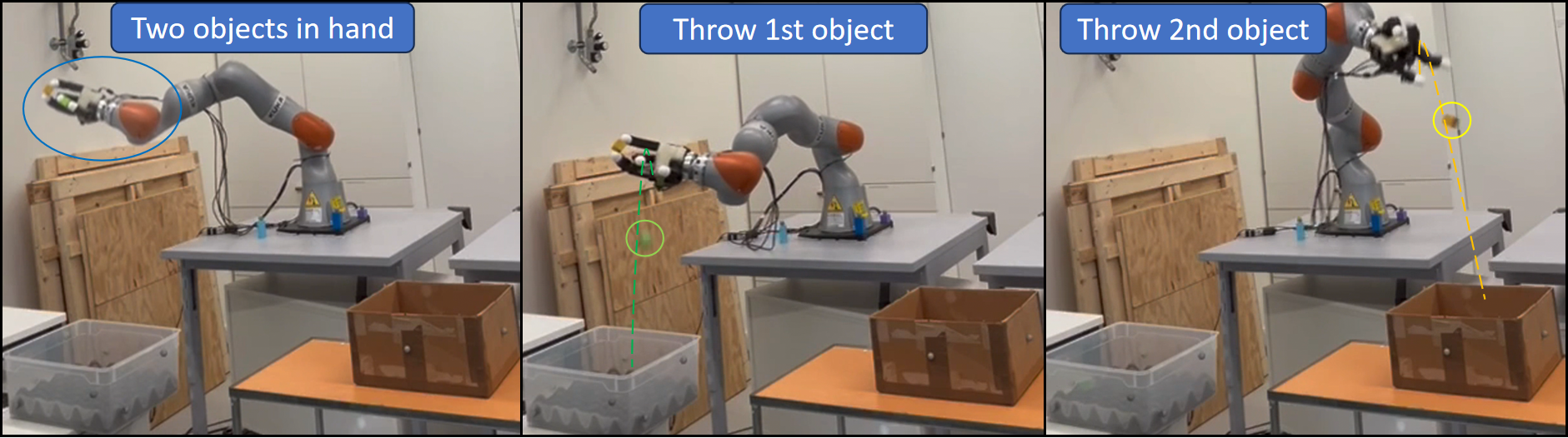}
    \caption{\small Multi-object multi-target throwing with a Kuka iiwa 7 robot equipped with an Allegro hand. The robot grasps two objects simultaneously and throws them to different target boxes in a single continuous motion, demonstrating the efficiency of our unified planning framework.}
    \label{fig:teaser}
\end{figure}
\textbf{List of contributions:}
\begin{itemize}
    \item We introduce a unified planning framework for multi-object multi-target throwing that combines explicit joint-space release state modeling with offline robot kinematic analysis. By exploiting rotational invariance of fixed-base manipulators, we reduce the dimensionality of the feasibility problem, enabling the generation of diverse feasible throwing configurations in under 5~ms.
    \item We propose a decomposition approach that separates single-target solution generation from multi-target coordination, and introduce a pruned Greedy search that retains only the top-$K$ candidates per target ranked by joint-space proximity to the current robot state. This enables practitioners to navigate the trade-off between planning time and execution optimality based on application requirements.
    \item We provide the first real-world demonstration of MOMT throwing, showing reactive multi-target adaptation and a 29\% reduction in execution time compared to independent single-object planning, and demonstrating reactive re-planning under dynamic target changes.
\end{itemize}

\section{Related work}
\subsection{Robot Throwing Planning}
Over the past decades, there has been tremendous progress on robot throwing~\cite{aboaf1987task, lynch1999dynamic, senoo2008high, zhang2012sampling, pekarovskiy2013optimal,zeng2020tossingbot, 
liu2022solution, kasaei2023throwing, liu2024tube, zermane2024planning, werner2024dynamic, ma2025learning}. Existing robot throwing motion synthesis methods can be grouped according to how the release state is specified: a) Implicit, b) Explicit Cartesian-space, c) Explicit joint-space.

\paragraph{Implicit release state}
Implicit formulations jointly determine the release state and trajectory. While this unifies optimization and learning-based methods, it typically results in difficult nonconvex optimization problems~\cite{lombai2009throwing} or sparse-reward MDPs that are hard to train efficiently~\cite{munn2024whole}.

\paragraph{Explicit Cartesian release state} This approach defines the release state in Cartesian space before synthesizing joint motion, leveraging efficient flight dynamics solvers. However, the selected Cartesian state may prove kinematically infeasible, necessitating costly replanning~\cite{zhang2012sampling} or manual tuning to ensure convergence~\cite{ma2025learning}.

\paragraph{Explicit joint-space formulation}
The explicit joint-space formulation specifies the release configuration directly in the robot’s joint state (position and velocity).~\cite{liu2022solution} proposed a planning framework for mobile manipulator throwing, enabling the generation of diverse throwing configurations in joint space in real time. However,~\cite{liu2022solution} assumes that the base is omnidirectional, and hence the horizontal invariance greatly simplifies the planning problem.
In this work, we adopt this explicit joint-space formulation and extend it to fixed-base manipulator throwing, and provide efficient planning solutions for multi-object multi-target scenarios. These two challenges were not addressed in previous works.

\subsection{Multi-object manipulation}
Holding and manipulating multiple objects simultaneously with a dexterous hand has long been recognized as important for improving robotic throughput. Early work by~\cite{oztop2006dexterous} demonstrated multi-object in-hand manipulation on a 16-DoF robotic hand through human skill transfer, establishing the feasibility of dexterous multi-object control. On the planning side,~\cite{yao2023exploiting} proposed a grasp synthesis framework that exploits the kinematic redundancy of a multi-fingered hand to generate stable multi-object grasps. More recently,~\cite{li2024multigrasp} introduced a learning-based pipeline for simultaneously grasping and lifting multiple objects with a dexterous hand, and~\cite{gao2026detachable} presented a hand design with reversible fingers that simplifies multi-object grasp planning.

Despite this progress, all existing multi-object manipulation work focuses on quasi-static regimes---grasping, holding, and repositioning. None addresses multi-object \emph{dynamic} manipulation, where objects must be launched at high speed toward distinct targets. In this work, we assume that a suitable multi-object grasp has already been planned (e.g., using~\cite{yao2023exploiting, li2024multigrasp}) and that the objects are placed correctly in the hand prior to throwing. Under this assumption, our framework addresses the open problem of planning coordinated throwing motions for multiple objects held simultaneously.
\section{Method}

Our framework consists of two components: (1) a \textbf{single-target throwing planner} that generates diverse feasible throwing configurations for a fixed-base manipulator in under 5~ms, and (2) a \textbf{multi-target coordinator} that coordinates multiple throws to minimize total task time. We describe each in turn.

\subsection{Assumptions}
\label{sec:assumptions}
We make the following assumptions throughout this work:
\begin{enumerate}
\item The manipulator is fixed-base, with the first joint rotating about the vertical axis (common in industrial arms).
\item Objects are point masses during flight (no aerodynamic effects or spin).
\item Grasp configurations are predefined: each object's position relative to the end-effector at release is known.
\item The workspace is obstacle-free during the throwing motion.
\item Object release is instantaneous, while in-hand sliding/pivoting are neglected.
\end{enumerate}

\subsection{Single-target Throwing Planner}

\subsubsection{Key Insight}
The single-target throwing planner builds upon the spatially-discretized dictionary, \emph{Velocity Hedgehog}, proposed in~\cite{liu2022solution} for mobile manipulator throwing. Counterintuitively, fixed-base throwing is \emph{more challenging} than mobile-base throwing: the mobile base provides additional degrees of freedom that actually \emph{simplify} the planning problem. Specifically,~\cite{liu2022solution} assumes an omnidirectional mobile base, where the base position can be determined \emph{after} the arm motion is planned—the horizontal invariance decouples the problem.

For fixed-base manipulators, this decoupling is lost. A straightforward extension would require discretizing the end-effector position $(x, y, z)$ along with throwing direction $(\phi, \gamma)$, resulting in a 5-dimensional mesh that is computationally intractable.

We exploit the fact that for common manipulators where the first joint rotates around the vertical axis, the throwing problem exhibits \emph{rotational invariance}. This allows us to:
\begin{itemize}
    \item Replace 2D horizontal position $(x,y)$ with 1D radial distance $d$, reducing to a 4D mesh $(z, d, \phi, \gamma)$.
    \item Compute yaw angle $\phi$ \emph{analytically} from the target position, achieving zero discretization error in landing position.
\end{itemize}
The result is a compact dictionary with effectively 4 discretized dimensions, which we call the \textbf{maximum velocity dictionary}. Combined with the \textbf{Backward Reachable Tube (BRT)} encoding valid release states from flight dynamics, the single-target throwing planner generates feasible throwing configurations in under 5~ms.

It is worth noting that this reduction relies on Assumption~1; for general manipulator mountings whose first joint is not vertical, rotational invariance no longer holds and one must discretize the full end-effector position $(x, y, z)$, enlarging the offline dictionary, while the online planning procedure remains unchanged.

\subsubsection{Geometric Modeling}
\begin{figure}[t!]
    \centering
    \includegraphics[width=0.75\linewidth]{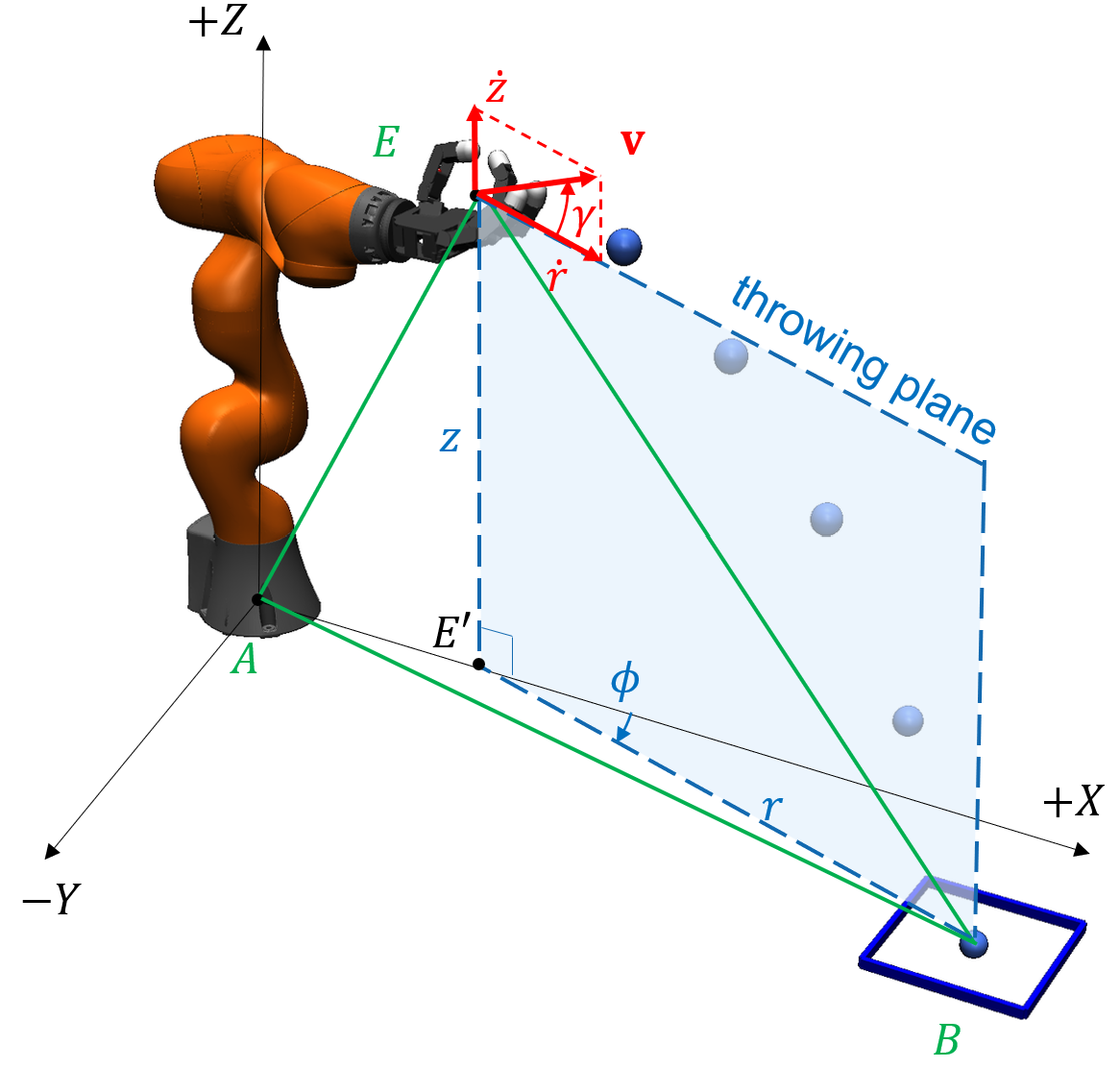}
    \caption{\small Geometric modeling at release.}
    \label{fig:geometric_modeling}
\end{figure}

We define the throwing geometry using triangle $\Delta AEB$ in 3D space (Fig.~\ref{fig:geometric_modeling}, Table~\ref{tab:symbols}). 

\begin{table}[h]
\centering
\renewcommand{\arraystretch}{1.2}
\begin{tabular}{clc}
\hline
\textbf{Symbol} & \textbf{Description} \\
\hline
$(q, \dot{q})\in \mathbb{R}^{2n}$ & Joint release state for n-DoF robot  \\ 
$^Ap^B \in \mathbb{R}^3$ & Target position in robot frame\\
$^Ap^E \in \mathbb{R}^3$ & Object position at release in robot frame \\
$\mathbf{v} \in \mathbb{R}^3$  & Object velocity at release  \\
$d_E$ & Horizontal distance from $E$ to robot base $A$  \\
$d_B$ & Horizontal distance from $B$ to robot base $A$  \\
$r$ & Horizontal distance from $E$ to $B$ \\
$z$ & Height of object relative to target  \\
$\phi$ & Yaw angle (throwing direction)  \\
$\gamma$ & Pitch angle (throwing elevation)  \\
$\psi_E, \psi_B$ & Azimuth of end-effector and target \\
\hline
\end{tabular}
\caption{\small Notation for geometric modeling}
\label{tab:symbols}
\end{table}

The throwing triangle parameters are computed via trigonometry (Fig.~\ref{fig:geometric_modeling}):
\begin{align}
r^2 &= d^2 + d_B^2 - 2 d d_B \cos(\psi_B - \psi_E) \label{eq:r_trig}\\
\phi &= \arctan\left(\frac{d_B \sin(\psi_B - \psi_E)}{d_B \cos(\psi_B - \psi_E) - d}\right) \label{eq:phi_trig}
\end{align}
Given target position $B$ relative to the robot base $A$, we seek a release state $(r, z, \dot{r}, \dot{z})$ satisfying joint limits such that the object lands on the target under gravity.



Thus, the throwing problem is formulated as:
\begin{equation}
\text{Find} \quad \{r, z, \dot{r}, \dot{z}, q, \dot{q}\}
\label{eq:throwing_problem}
\end{equation}
subject to:
\begin{align}
    \mathbf{v} &= J(q)\dot{q} \label{eq:vel} \\
    \frac{\mathbf{v}_{xy}}{\|\mathbf{v}_{xy}\|} &= \frac{^Ep^B_{xy}}{\|{^Ep^B_{xy}}\|} \label{eq:align}\\
    (r(q),\, z(q),\, \|\mathbf{v}_{xy}\|,\, \mathbf{v}_z) &\in \mathrm{BRT}_{f_{fly}} \label{eq:brt} \\
q_{\min} \leq q \leq q_{\max}&, \quad \dot{q}_{\min} \leq \dot{q} \leq \dot{q}_{\max}
\end{align}
where $r$ and $z$ depend on $q$ implicitly via forward kinematics and the throwing 
geometry (Fig.~\ref{fig:geometric_modeling}); \eqref{eq:align} enforces that the horizontal 
release velocity points toward the target, constraining the throw to lie in plane $EE'B$; 
and $\mathrm{BRT}_{f_{\text{fly}},\mathcal{X}}$ is the set of valid release states in that 
plane, which we explain next.

\subsubsection{Backward Reachable Tube (BRT)}
\begin{figure}[t]
    \centering
    \includegraphics[width=0.8\linewidth]{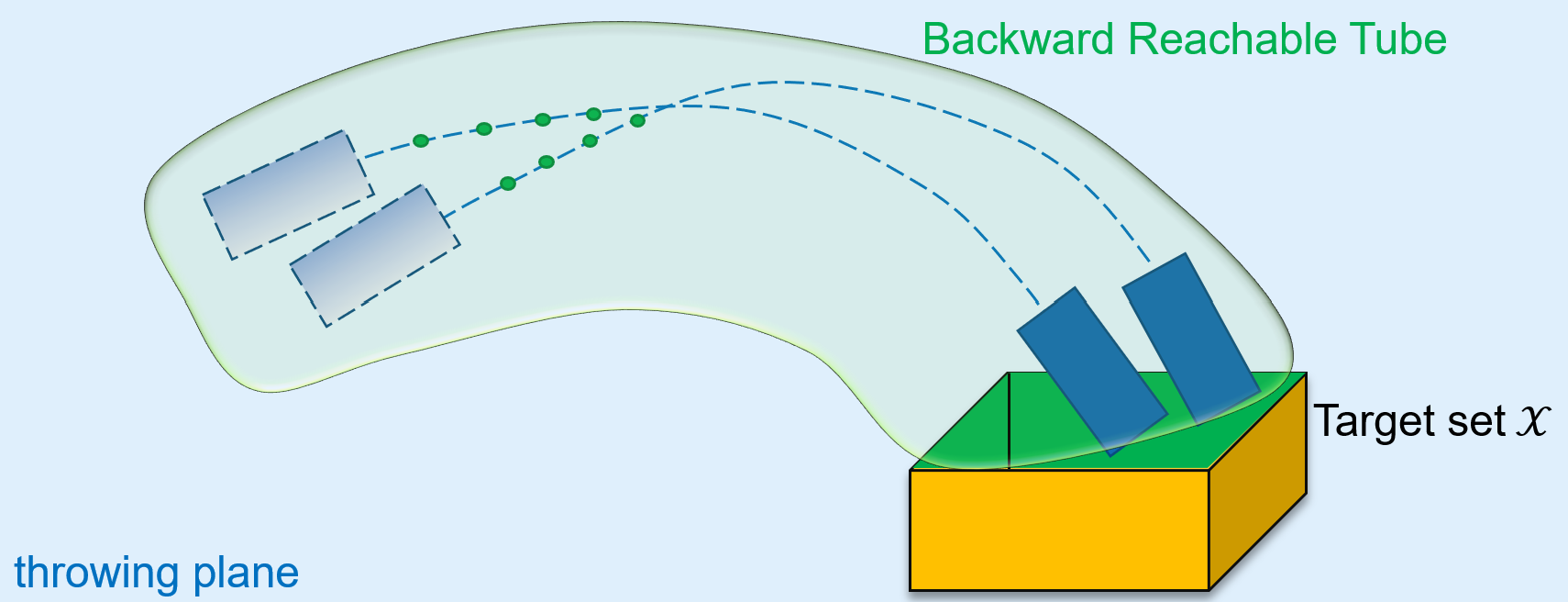}
    \caption{\small The Backward Reachable Tube (BRT) is the set of release states whose trajectories reach the landing target set $\mathcal{X}$.}
    \label{fig:vis-brt_traj}
\end{figure}

The BRT encodes all release states from which an object can reach the target. We define the flight state as $\xi = [r, z, \dot{r}, \dot{z}]^\top \in \mathbb{R}^4$, governed by: $\dot{\xi} = f_{\text{fly}}(\xi)$, where $f_{\text{fly}}$ describes the flight dynamics (e.g., projectile motion under gravity). Let $\zeta_{f_{\text{fly}},\xi^0}(t)$ denote the trajectory starting from state $\xi^0$. Given a landing target set $\mathcal{X} \subseteq \mathbb{R}^4$ specifying allowable landing positions and velocities, the BRT is:
\begin{equation}
\text{BRT}_{f_{\text{fly}},\mathcal{X}} = \{\xi^0 \mid \exists\, t \geq 0,\; \zeta_{f_{\text{fly}},\xi^0}(t) \in \mathcal{X}\}
\end{equation}
In words: the BRT contains all states from which the object eventually lands in $\mathcal{X}$ (Fig.~\ref{fig:vis-brt_traj}).

\textbf{Building the BRT:} We sample states in $\mathcal{X}$ and integrate the ODE \emph{backwards} in time to trace trajectories to their release states. For projectile motion (gravity only), the flight time and landing position have closed-form expressions:
\begin{equation}
\Delta t_{\text{fly}} = \frac{\dot{z}_d + \sqrt{\dot{z}_d^2 + 2g(z_d - z_{\text{land}})}}{g}, \quad
r_{\text{land}} = r_d + \dot{r}_d \Delta t_{\text{fly}}
\end{equation}
where $(r_d, z_d, \dot{r}_d, \dot{z}_d)$ is the release state. For complex dynamics (e.g., with air drag~\cite{lubarda2022review}), numerical integration can be used.

\begin{figure}[t]
    \centering
    \includegraphics[width=0.95\linewidth]{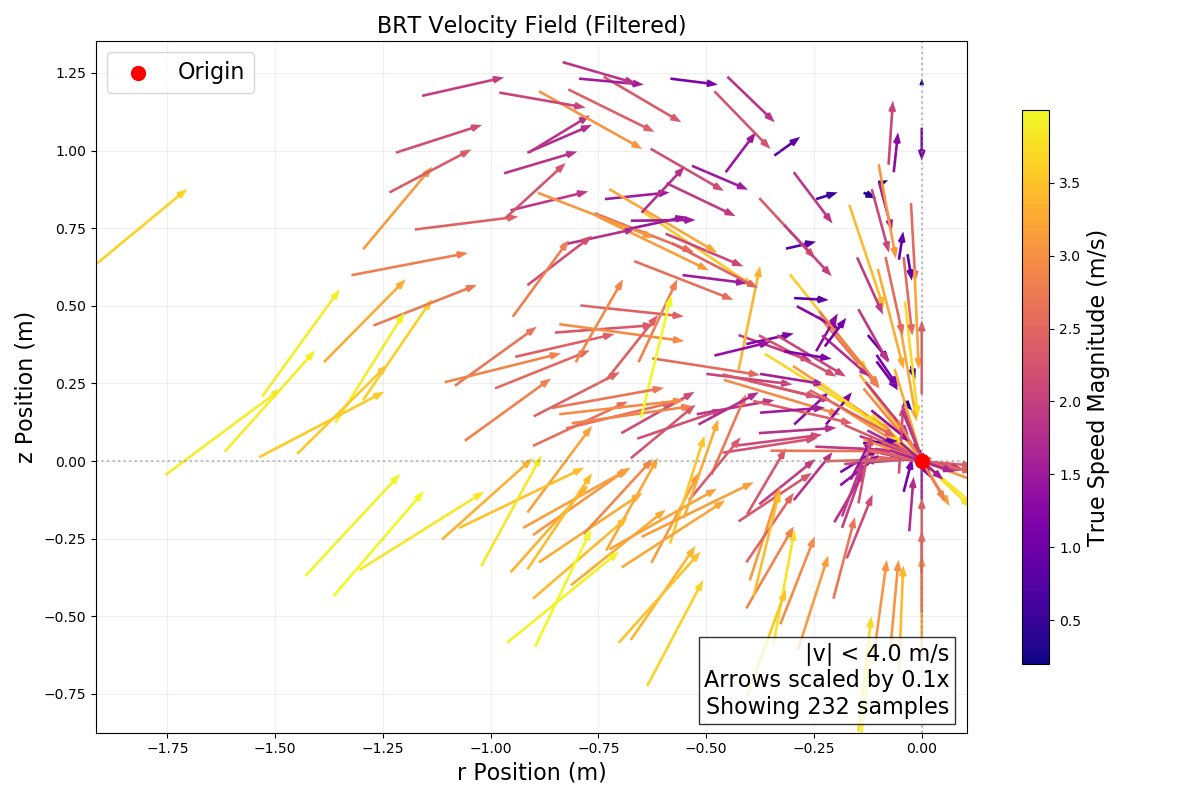}
    \caption{\small Visualization of the BRT. At each spatial position, arrows show valid release velocities.}
    \label{fig:vis-brt}
\end{figure}

The resulting BRT is a 4D volume in state space: at each position $(r, z)$, there exists a \emph{set} of valid release velocities $(\dot{r}, \dot{z})$ that will land the object in the target (Fig.~\ref{fig:vis-brt}). This multiplicity of solutions is key to our approach—it provides flexibility for matching with robot kinematic constraints.

\subsubsection{Maximum Velocity Dictionary}
The maximum velocity dictionary stores the robot's maximum achievable end-effector velocity for each throwing configuration, organized as a 4D mesh over $(z, d, \phi, \gamma)$.

Since the first joint $q_0$ rotates around the vertical axis, we exclude it during offline construction (setting $q_0 = 0$). During online matching, $q_0$ is computed analytically to align the throwing triangle with the target azimuth:
\begin{equation}
q_0 = \psi_B - \arccos\left(\frac{d^2 + d_B^2 - r^2}{2 d d_B}\right)
\label{eq:q0_computation}
\end{equation}
This yields zero discretization error in the horizontal landing position. Therefore, for each bin, we store (Algorithm~\ref{alg:MVD}):
\begin{itemize}
    \item Maximum feasible velocity $v_{\max}(z, d, \phi, \gamma)$
    \item Corresponding joint configuration $q$
\end{itemize}

The maximum velocity at configuration $q$ along direction $(\phi, \gamma)$ is found by solving:
\begin{equation}
\max_s \; s \quad \text{s.t.} \quad \dot{q}_{\min} \leq J^{\dagger}(q) \vec{v} \leq \dot{q}_{\max}
\end{equation}
where $\vec{v} = s[\cos\gamma\cos\phi,\; \cos\gamma\sin\phi,\; \sin\gamma]^\top$ and $J^{\dagger}(q)$ is the Jacobian pseudoinverse. This is a linear program (LP) solved efficiently as a subroutine.
\begin{figure}[h!]
    \centering
    \includegraphics[width=0.95\linewidth]{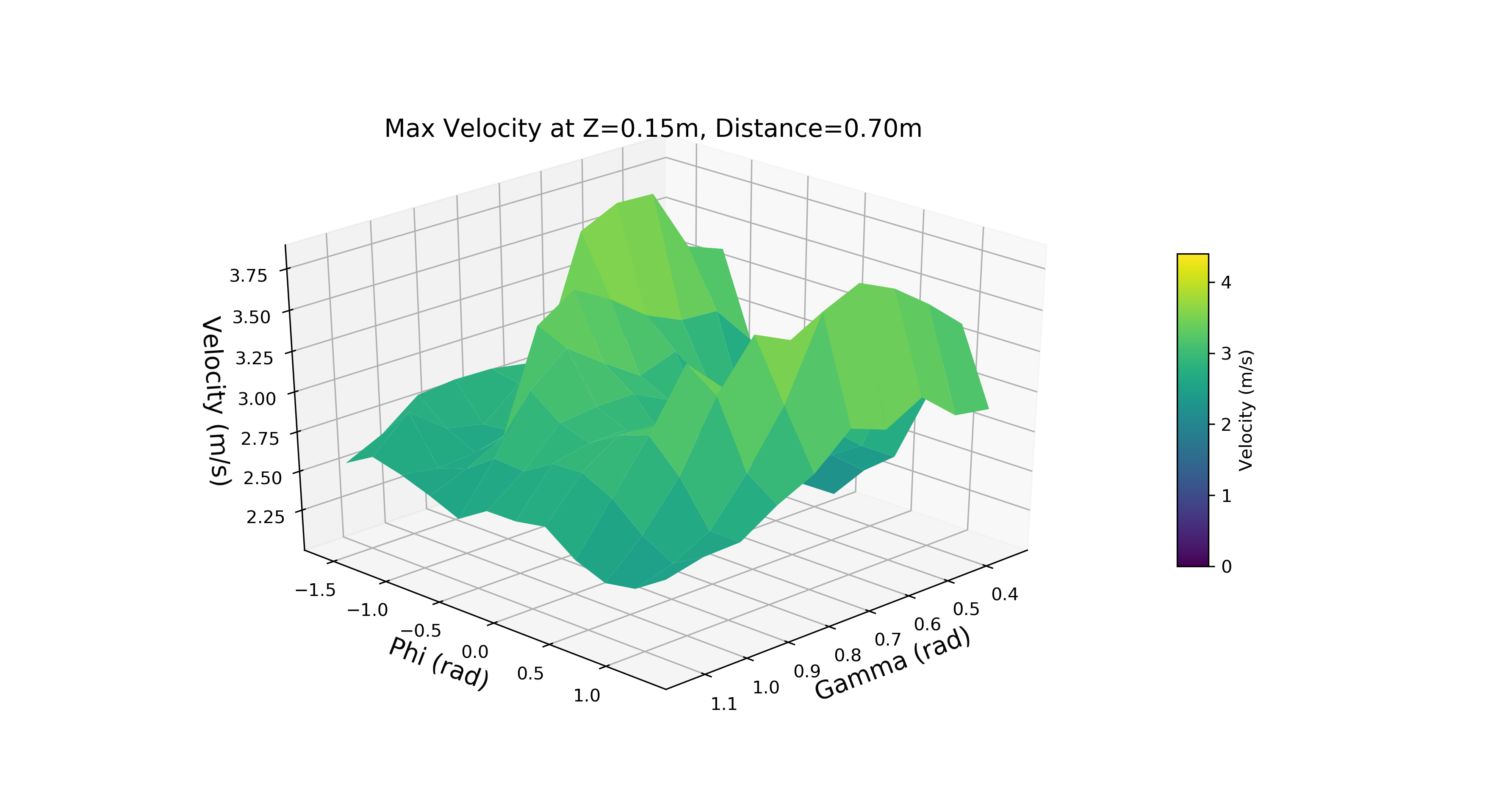}
    \caption{\small Visualization of the maximum velocity dictionary for the Kuka iiwa 7. The surface shows the maximum achievable end-effector velocity as a function of throwing direction $(\phi, \gamma)$ at a fixed height and distance.}
    \label{fig:vis-mvd}
\end{figure}

Fig.~\ref{fig:vis-mvd} visualizes the maximum velocity dictionary for the Kuka iiwa 7. The distribution reveals kinematic limits across configurations: for this robot, the maximum Cartesian velocity is $\sim$3.75~m/s, and the maximum throwing distance (determined by flight dynamics) is $\sim$1.5~m.

\begin{algorithm}[t]
\caption{Maximum Velocity Dictionary \textbf{(Offline)}}
\label{alg:MVD}
\begin{algorithmic}[1]
\STATE \textbf{Input:} Robot model, joint position and velocity limits $(q_{\min}, q_{\max}, \dot{q}_{\min}, \dot{q}_{\max})$
\STATE \textbf{Output:} $v_{\max}(z,d,\phi,\gamma)$, $q(z,d,\phi,\gamma)$
\STATE Sample joint configurations $Q$ on a grid
\STATE Filter configurations with small Jacobian singular values
\STATE Group by height $z$ and distance $d$
\FOR{each $(z, d, \phi, \gamma)$ bin}
    \FOR{each configuration $q$ in bin}
        \STATE $v \gets$ solve LP for max velocity along $(\phi, \gamma)$
        \IF{$v > v_{\max}(z, d, \phi, \gamma)$}
            \STATE $v_{\max}(z, d, \phi, \gamma) \gets v$; \quad $q(z,d,\phi,\gamma) \gets q$
        \ENDIF
    \ENDFOR
\ENDFOR
\end{algorithmic}
\end{algorithm}

\subsubsection{Solution Matching}
Given a target position, we find feasible throwing configurations by matching the BRT (valid release states) with the maximum velocity dictionary (achievable robot states).

A release state $\xi = (r, z, \dot{r}, \dot{z})$ from the BRT is feasible if the required velocity $v = \|(\dot{r}, \dot{z})\|$ does not exceed the robot's capability:
\begin{equation}
v \leq v_{\max}(z, d, \phi, \gamma), \quad \text{where } \gamma = \arctan(\dot{r}/\dot{z})
\end{equation}

\textbf{Efficient matching:} For each height $z$ and pitch $\gamma$, we have two sorted sets: BRT states by required velocity and maximum velocity dictionary entries by maximum velocity. Matching reduces to a linear sweep with complexity $\mathcal{O}(N_{\text{BRT}} + N_{\text{dict}})$ instead of naive $\mathcal{O}(N_{\text{BRT}} \cdot N_{\text{dict}})$.

After matching, we filter candidates based on task-specific criteria (e.g., prefer configurations with safety margins) and output the final throwing configuration $(q, \dot{q})$.

\subsubsection{Trajectory Generation}
Given the current robot state and a candidate throwing configuration from Solution Matching, we use Ruckig~\cite{berscheid2021jerk} to generate time-optimal jerk-limited trajectories. Ruckig provides both the executable 
trajectory and its duration in closed form ($<0.5$~ms for 7-DoF). This trajectory generation is also used in the multi-target coordinator to evaluate and compare total execution time across multi-throw sequences.

\subsection{Multi-target Coordinator}

\begin{figure}[t]
    \centering
    \includegraphics[width=0.85\linewidth]{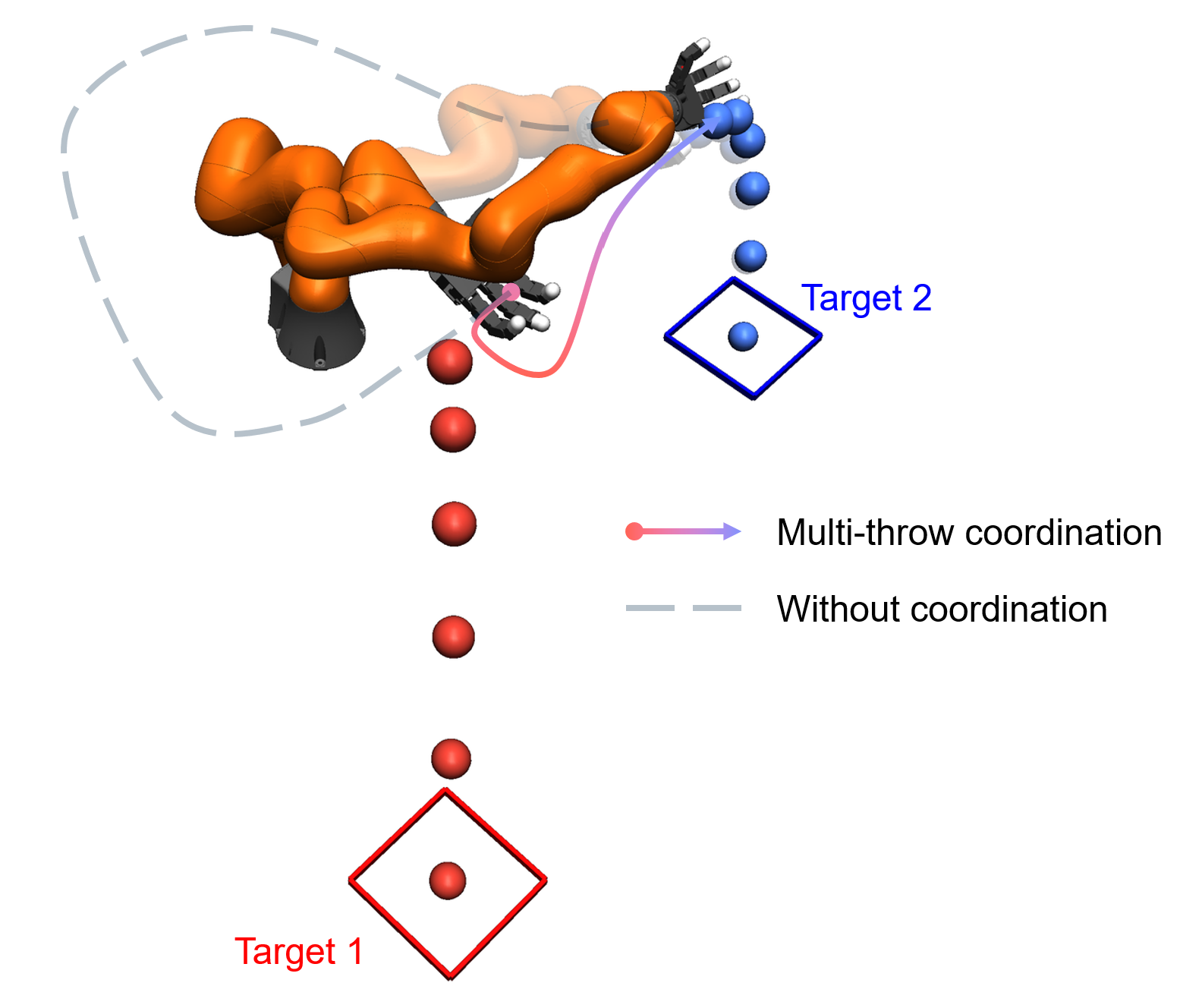}
    \caption{\small Motivation for multi-throw coordination. The single-target planner yields multiple valid configurations per target (red for Target~1, blue for Target~2). Choosing configurations independently may result in large joint-space transitions (dashed arc), while coordinated selection minimizes total execution time (solid arrow).}
    \label{fig:coordination_motivation}
\end{figure}

The single-target throwing planner outputs a \emph{set} of valid throwing configurations per target. When multiple targets must be served, the choice of which configuration to use for each target is not independent: two configurations that are individually valid may be far apart in joint space (Fig.~\ref{fig:coordination_motivation}, dashed arc), leading to long transition trajectories. The multi-target coordinator selects one configuration per target \emph{jointly}, minimizing total execution time (solid arrow).

\begin{figure}[ht]
    \centering
    \includegraphics[width=0.8\linewidth]{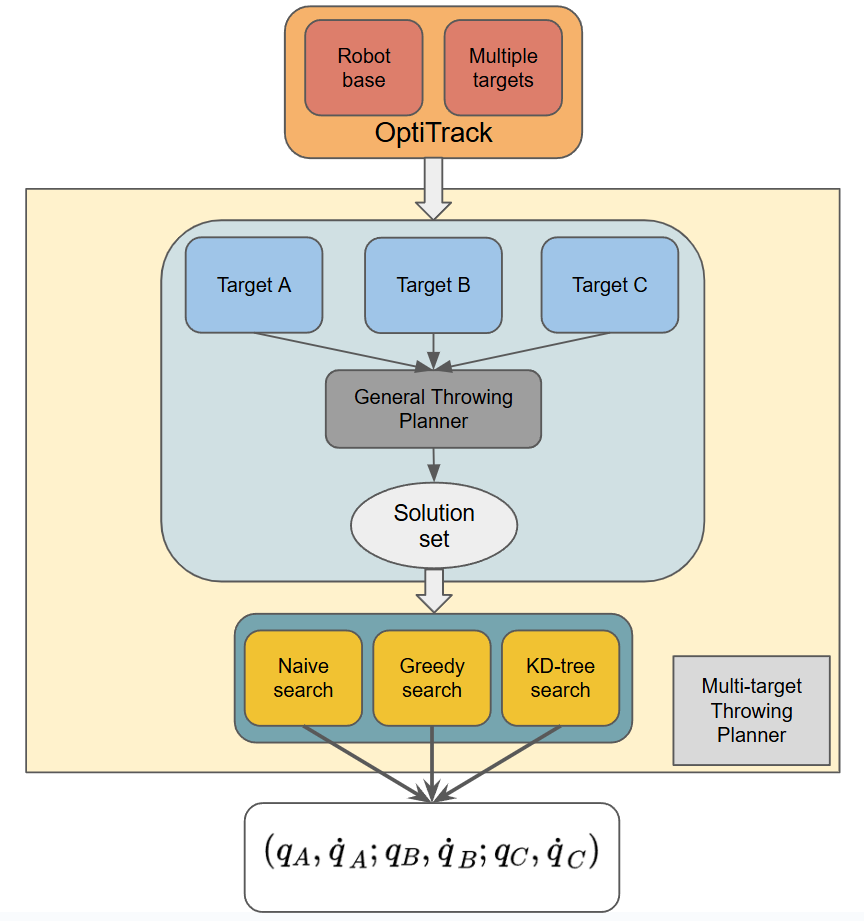}
    \caption{\small Multi-target coordinator: Given multiple targets, the single-target throwing planner generates candidate solutions for each. The coordinator searches for the best combination minimizing total execution time.}
    \label{fig:MTS}
\end{figure}

For multi-object multi-target (MOMT) throwing, we must coordinate multiple throws to minimize total task time. This introduces a combinatorial challenge: with $N$ objects and $M$ candidate configurations per object, there are $M^N$ possible combinations. We consider three search strategies (Fig.~\ref{fig:MTS}) trading off computation time vs.\ execution time:

\subsubsection{Naive Search}
Solve each target independently using single-target throwing planner, selecting the locally optimal configuration for each. This ignores inter-throw coordination, leading to suboptimal total execution time (e.g., large joint motions between consecutive throws).

\begin{algorithm}[h]
\caption{Naive Search \textbf{(Online)}}\label{alg:naive_search}
\begin{algorithmic}[1]
\STATE \textbf{Input:} Targets $\mathcal{T} = \{t_1, \ldots, t_N\}$, robot base $A$
\STATE \textbf{Output:} Trajectory sequence
\FOR{each target $t_i \in \mathcal{T}$}
    \STATE $(q_i^*, \dot{q}_i^*) \gets \text{SingleTargetPlan}(t_i, A)$ \hfill \textcolor{blue}{// Local optimum}
\ENDFOR
\STATE \textbf{return} TrajectoryGenerator($\{(q_i^*, \dot{q}_i^*)\}$)
\end{algorithmic}
\end{algorithm}

\subsubsection{Greedy Search}
Generate multiple candidate configurations per target, then search over all combinations. This finds globally better solutions but has exponential complexity $\mathcal{O}(M^N)$.
\begin{algorithm}[h!]
\caption{Greedy Search \textbf{(Online)}}
\label{alg:greedy_search}
\begin{algorithmic}[1]
\STATE \textbf{Input:} Targets $\mathcal{T}$, robot base $A$
\STATE \textbf{Output:} Time-optimal trajectory sequence
\FOR{each target $t_i$}
    \STATE $\mathcal{C}_i \gets \text{GetCandidates}(t_i, A)$ \hfill \textcolor{blue}{// Multiple candidates}
\ENDFOR
\STATE $T_{\min} \gets \infty$
\FOR{each combination $(c_1, \ldots, c_N) \in \mathcal{C}_1 \times \cdots \times \mathcal{C}_N$}
    \STATE $T \gets$ total trajectory duration for this combination
    \IF{$T < T_{\min}$}
        \STATE $T_{\min} \gets T$; save best combination
    \ENDIF
\ENDFOR
\STATE \textbf{return} best trajectory sequence
\end{algorithmic}
\end{algorithm}

\subsubsection{Pruned Greedy Search}
To control computational cost, we introduce a pruned variant of Greedy search with parameter $K$: for each target, only the top-$K$ candidates ranked by the heuristic distance $\|q_c - q_0\|_2 + 
2\|\dot{q}_c - \dot{q}_0\|_2$, where $(q_0, \dot{q}_0)$ is the 
current robot state and $(q_c, \dot{q}_c)$ is a candidate 
configuration. The intuition is that configurations close to the 
current robot state in joint position and velocity require shorter 
transition trajectories, and are therefore more likely to yield 
low total execution time.


\section{Experiment}
In this section, we evaluate the proposed MOMT throwing framework through simulation studies and real-world experiments. Our evaluation investigates three key questions:
\begin{enumerate}
    \item \textbf{Planning-execution trade-off}: How does the pruning parameter $K$ in Greedy search affect the balance between computation time and execution optimality? 
    \item \textbf{Scalability}: How does the framework perform as the number of objects increases?
    \item \textbf{Real-world validity}: Is the planning framework effective on physical hardware, and can the computational efficiency enable reactive re-planning? 
\end{enumerate}

We first systematically study the planning-execution trade-off in simulation where extensive trials can be conducted efficiently (Sections~\ref{sec:sim_study}--\ref{sec:scalability}), then validate the framework on real hardware with reactive throwing demonstrations (Section~\ref{sec:real_world}).

\subsection{Simulation Study: Two-Object Throwing}
\label{sec:sim_study}
We characterize the planning-execution trade-off in simulation using a Kuka iiwa 7 model with the same kinematic parameters and joint limits as the real robot. All timing measurements were performed on Apple M2 Pro (3.49GHz, 32GB). 

Beyond computation and execution time, we also report energy cost, measured from the initial holding configuration to release of the last object (excluding the homing phase). Joint torques follow from the planned trajectory via rigid-body inverse dynamics:
\begin{equation}
    \boldsymbol{\tau}(t) = \mathbf{M}(\mathbf{q}_{\mathrm{des}})\ddot{\mathbf{q}}_{\mathrm{des}} + \mathbf{C}(\mathbf{q}_{\mathrm{des}}, \dot{\mathbf{q}}_{\mathrm{des}})\dot{\mathbf{q}}_{\mathrm{des}} + \mathbf{g}(\mathbf{q}_{\mathrm{des}})
    \label{eq:inverse_dynamics}
\end{equation}
where $\mathbf{M}$, $\mathbf{C}$, and $\mathbf{g}$ are the mass, Coriolis, and gravity terms. Since most manipulators dissipate braking energy as heat rather than recovering it, we count only positive mechanical work:
\begin{equation}
    E = \int_{t_0}^{t_{\mathrm{release}}} \max\!\bigl(\boldsymbol{\tau}(t) \cdot \dot{\mathbf{q}}_{\mathrm{des}}(t),\; 0\bigr) \, \mathrm{d}t
    \label{eq:energy}
\end{equation}
where $t_0$ is the start of the throwing motion and $t_{\mathrm{release}}$ is the time at which the last object is released.

\subsubsection{Trade-off Analysis}
Table~\ref{tab:tradeoff_2obj_sim} presents the trade-off between computation time and execution time for varying pruning levels $K$ in the two-object scenario ($n=10$ target configurations).
\begin{table}[h]
    \centering
    \caption{Planning-execution trade-off for two objects (simulation)}
    \label{tab:tradeoff_2obj_sim}
    \begin{tabular}{lccc}
        \hline
        Method & Computation (s) & Execution (s) & Energy (J) \\
        \hline
        Naive ($K=1$) & $0.06 \pm 0.01$ & $6.38 \pm 1.69$ & $87.46 \pm 58.74$ \\
        Greedy-$K5$ & $0.06 \pm 0.01$ & $5.10 \pm 1.08$ & $62.56 \pm 40.52$ \\
        Greedy-$K10$ & $0.06 \pm 0.01$ & $4.40 \pm 0.70$ & $53.56 \pm 36.68$ \\
        Greedy-$K20$ & $0.07 \pm 0.01$ & $4.21 \pm 0.68$ & $43.50 \pm 24.63$ \\
        Greedy-$K50$ & $0.12 \pm 0.01$ & $3.61 \pm 0.41$ & $42.39 \pm 20.37$ \\
        Greedy-$K80$ & $0.19 \pm 0.01$ & $3.51 \pm 0.25$ & $47.38 \pm 21.90$ \\
        Greedy-$K100$ & $0.27 \pm 0.01$ & $3.47 \pm 0.23$ & $47.92 \pm 14.12$ \\
        Greedy (full) & $0.55 \pm 0.11$ & $3.45 \pm 0.25$ & $46.27 \pm 19.90$ \\
        \hline
    \end{tabular}
\end{table}
The results reveal:
\begin{itemize}
    \item \textbf{Diminishing returns}: Increasing $K$ beyond 50 yields marginal execution improvement while computation time continues to grow.
    \item \textbf{Sweet spot}: At $K=80$, the pruned Greedy achieves 98.3~\% of the full Greedy's execution performance while requiring only 0.19~s of computation.
\end{itemize}

\subsection{Scalability and Planning-Execution Trade-off for More Objects}
\label{sec:scalability}

To evaluate the scalability of our framework, we extended the simulation to scenarios with three and four objects ($n=10$ target configurations each). Table~\ref{tab:tradeoff_3obj_sim2} and Table~\ref{tab:tradeoff_4obj_sim} present the planning-execution trade-offs.

\begin{table}[h]
    \centering
    \caption{Planning-execution trade-off for three objects (simulation)}
    \label{tab:tradeoff_3obj_sim2}
    \begin{tabular}{lccc}
        \hline
        Method & Computation (s) & Execution (s) & Energy (J) \\
        \hline
        Naive ($K=1$) & $0.11 \pm 0.01$ & $9.57 \pm 1.58$ & $128.74 \pm 74.57$ \\
        Greedy-$K5$ & $0.10 \pm 0.01$ & $6.98 \pm 0.84$ & $73.08 \pm 34.82$ \\
        Greedy-$K10$ & $0.12 \pm 0.01$ & $6.41 \pm 0.56$ & $73.42 \pm 23.51$ \\
        Greedy-$K20$ & $0.29 \pm 0.01$ & $5.85 \pm 0.33$ & $62.78 \pm 19.99$ \\
        Greedy-$K50$ & $2.81 \pm 0.06$ & $5.45 \pm 0.26$ & $67.35 \pm 18.84$ \\
        Greedy-$K80$ & $11.11 \pm 0.22$ & $5.27 \pm 0.21$ & $76.26 \pm 21.03$ \\
        Greedy-$K100$ & $21.81 \pm 0.38$ & $5.22 \pm 0.17$ & $67.86 \pm 27.24$ \\
        Greedy (full) & $68.13 \pm 13.85$ & $5.18 \pm 0.22$ & $78.41 \pm 30.56$ \\
        \hline
    \end{tabular}
\end{table}
\begin{table}[h]
    \centering
    \caption{Planning-execution trade-off for four objects (simulation)}
    \label{tab:tradeoff_4obj_sim}
    \begin{tabular}{lccc}
        \hline
        Method & Computation (s) & Execution (s) & Energy (J) \\
        \hline
        Naive ($K=1$) & $0.15 \pm 0.01$ & $13.59 \pm 1.96$ & $195.11 \pm 66.86$ \\
        Greedy-$K5$ & $0.15 \pm 0.01$ & $9.54 \pm 1.05$ & $95.19 \pm 41.76$ \\
        Greedy-$K10$ & $0.40 \pm 0.01$ & $8.95 \pm 0.82$ & $94.79 \pm 48.02$ \\
        Greedy-$K20$ & $3.90 \pm 0.08$ & $8.16 \pm 0.41$ & $100.74 \pm 21.65$ \\
        Greedy-$K30$ & $18.50 \pm 0.22$ & $7.73 \pm 0.25$ & $101.50 \pm 27.27$ \\
        Greedy-$K50$ & $138.45 \pm 1.35$ & $7.48 \pm 0.20$ & $87.10 \pm 33.74$ \\
        \hline
    \end{tabular}
\end{table}

Key observations from the multi-object experiments ($N=3, 4$) include:
\begin{itemize}
    \item \textbf{Scalability and Pruning Necessity}: The computational cost grows exponentially with object count ($O(K^N)$), rendering exhaustive search intractable (e.g., 68.1~s for $N=3$, $>138$~s for $N=4$ with $K=50$). However, the pruned \textbf{Greedy-$K20$} strategy effectively mitigates this, maintaining tractable computation ($0.29$~s for $N=3$, $3.9$~s for $N=4$) while capturing over \textbf{85\%} of the execution time improvement compared to the Naive baseline.
    \item \textbf{Consistent Coordination Benefits}: The framework demonstrates robust performance across complexities. The execution time reduction from Naive to the best Greedy solution remains stable at approximately \textbf{45\%} for both 3-object ($45.8\%$) and 4-object ($44.9\%$) scenarios, indicating that temporal coordination benefits do not diminish with increased task size.
    \item \textbf{Energy Efficiency}: Coordination significantly optimizes energy usage alongside time. Particularly in the 4-object case, the Naive approach consumes nearly double the energy ($195.11$~J) compared to Greedy variants ($\approx 95$--$100$~J), as the optimized sequence minimizes unnecessary holding torques and inefficient transitions.
\end{itemize}

\subsection{Real-World Validation}
\label{sec:real_world}

Having characterized the planning-execution trade-off in simulation, we then deployed and validated the framework on our available lab hardware using a two-object throwing task, confirming its real-world viability.

\subsubsection{Experimental Setup}
The real-world experiments were conducted using a Kuka iiwa 7 robot equipped with an Allegro hand. An OptiTrack motion capture system provided ground-truth measurements of object flight trajectories, target positions, and the robot base frame. The targets were boxes of $0.3\times0.3$~m. The planned throwing trajectories are tracked by a joint impedance controller.

\begin{figure}[h!]
    \centering
    \begin{minipage}[b]{0.45\linewidth}
        \centering
        \includegraphics[width=3cm, height=3cm]{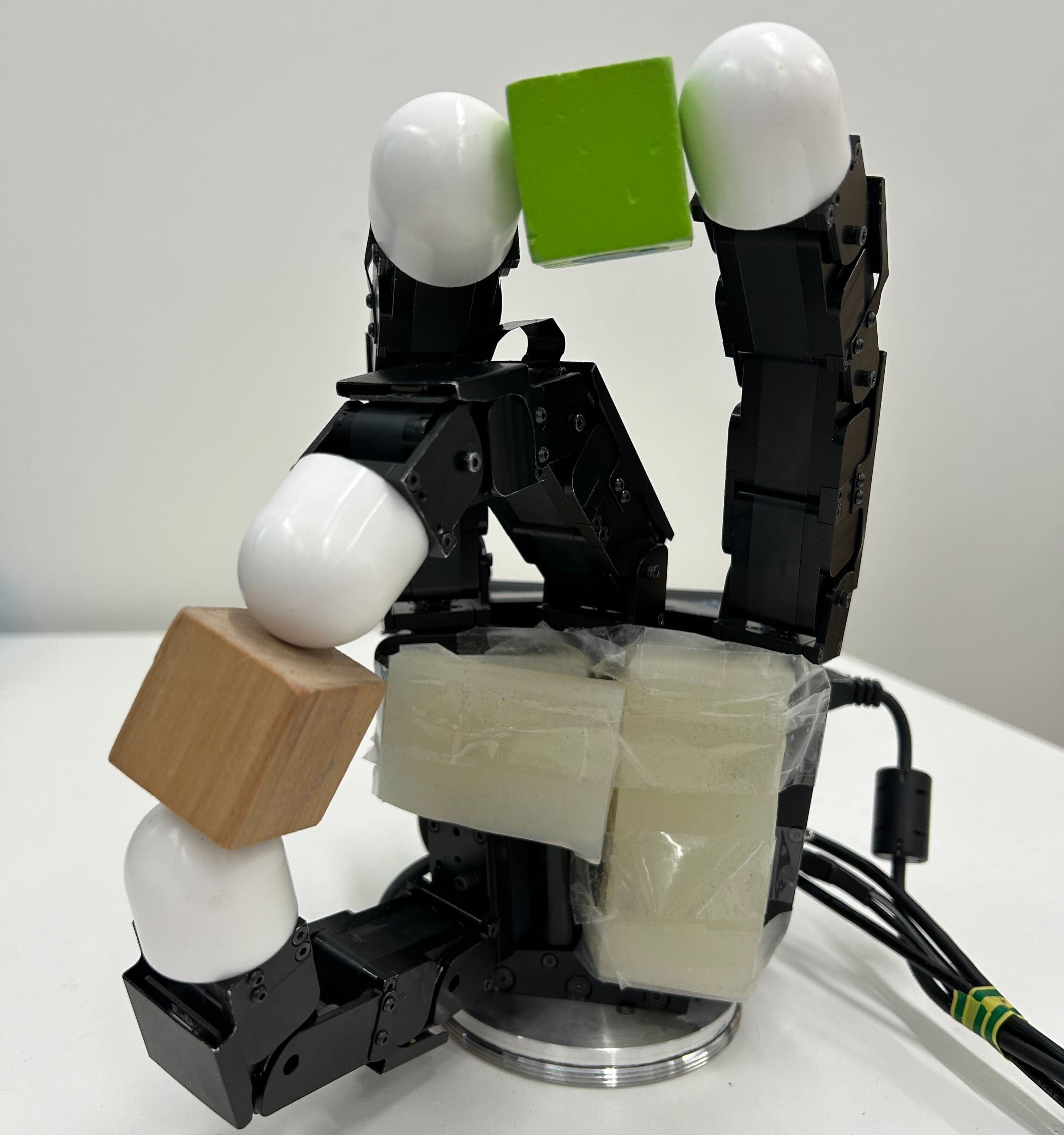}
        \label{fig:Grasping-a}
    \end{minipage}
    \begin{minipage}[b]{0.45\linewidth}
        \centering
        \includegraphics[width=3cm, height=3cm]{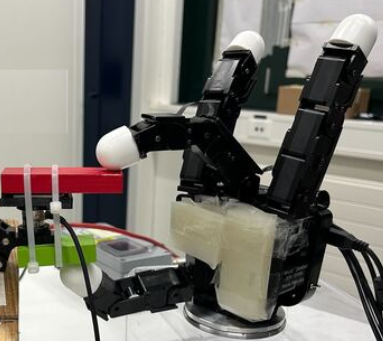}
        \label{fig:Grasping-b}
    \end{minipage}
    \caption{\small Left: Two objects held by fingertips in the Allegro hand using a manually-tuned configuration that minimizes release interference. Right: Release delay measurement using an ATI nano F/T sensor.}
    \label{fig:Grasping}
\end{figure}

The thrown objects were wooden blocks (mass: 15 g, dimensions: 3$\times$3$\times$3 cm). For multi-object grasping, we identified a fingertip configuration on the Allegro hand that can reliably hold two objects while minimizing release interference between them (Fig.~\ref{fig:Grasping}). Finding such configurations required manual tuning, as the Allegro hand's kinematics impose constraints on feasible multi-object grasps.

To account for the release delay inherent in finger opening, we command the hand to open before arriving at the nominal throwing state. The release delay was measured via an ATI nano F/T sensor (Fig.~\ref{fig:Grasping}) and found to be approximately 40-50~ms. This delay is compensated in the release timing.


\begin{table}[h]
    \centering
    \caption{Real-world validation (two objects, $n=10$ configurations)}
    \label{tab:real_validation}
    \begin{tabular}{lccc}
        \hline
        Method & Computation & Execution & Throws Landed \\
        \hline
        Naive & $0.06 \pm 0.02$~s & $5.20 \pm 0.98$~s & 14/20 \\
        Greedy & $69.8 \pm 32.3$~s & $3.69 \pm 0.37$~s & 14/20 \\
        \hline
    \end{tabular}
\end{table}

We first confirm that the planning-execution trade-off observed in simulation transfers to real hardware. Table~\ref{tab:real_validation} summarizes results across 10 target configurations. Consistent with simulation, Greedy search reduces execution time by 29\% compared to Naive search, at the cost of significantly longer planning time. The supplementary video shows both strategies side-by-side. 

Notably, both strategies land the same fraction of throws (14/20; Table~V). This is expected rather than incidental: the coordination strategy governs \emph{which} release states are selected and in what order---determining execution time and energy---but not the accuracy with which a chosen throw is realized. Landing accuracy instead depends on release dynamics, which are identical across both methods. The observed misses stem primarily from our simplified release model (Assumption~5): instantaneous release with neglected in-hand sliding and pivoting, compounded by release-timing variance and tracking error near the velocity limits. Improving landing accuracy is therefore orthogonal to our contribution and could be addressed by release-aware modeling and control---robustifying throws against release uncertainty for center-of-mass grasps \cite{liu2024tube}, modeling the transient release dynamics of eccentric grasps \cite{liu2025transient}, or refining throws from past experience \cite{liu2025learning}. These act at the level of individual throws and integrate directly with our planner.

\subsubsection{Reactive Multi-Target Throwing}

The key advantage of fast online planning is enabling reactive behavior. We demonstrate this by randomly changing target positions during execution.

\begin{figure*}[th!]
    \centering
    \includegraphics[width=1.0\linewidth]{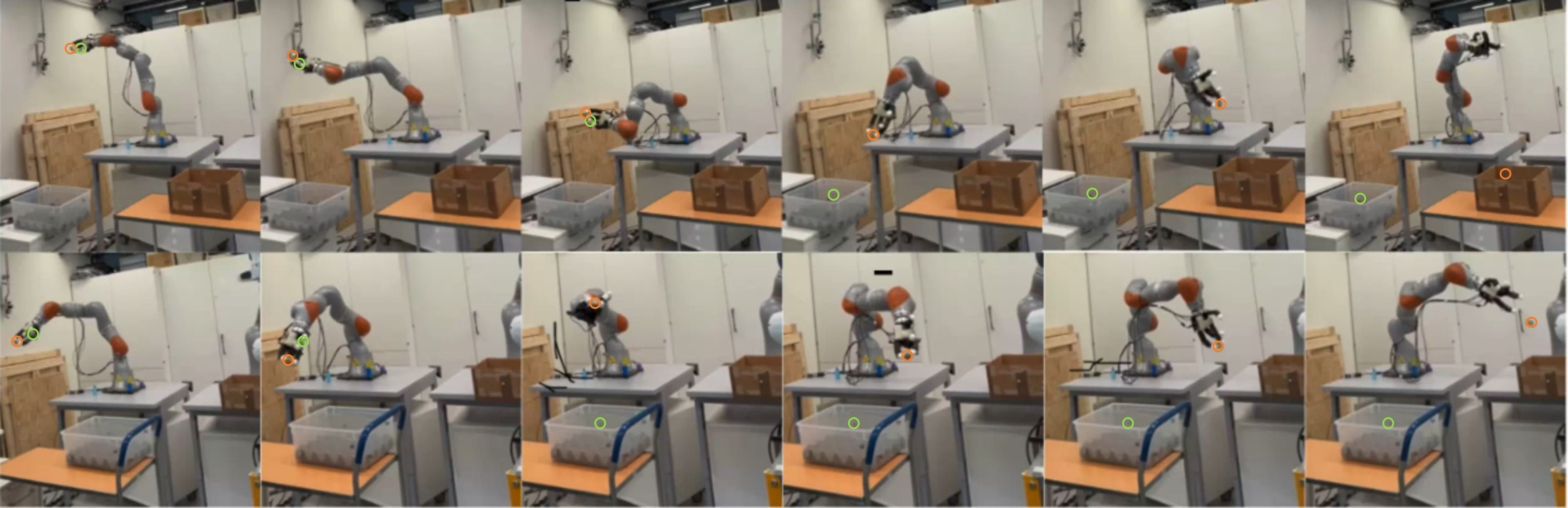}
    \caption{Reactive multi-object multi-target throwing. When target positions change mid-execution, the system re-plans and adapts in real-time. See the supplementary video for the full demonstration.} 
    \label{fig:reactive}
\end{figure*}

Figure~\ref{fig:reactive} illustrates the reactive throwing experiment. When a target position change is detected, the system:
\begin{enumerate}
    \item Queries the single-target planner for new feasible configurations for the updated target.
    \item Updates the multi-target coordination solution if the robot has not yet committed to the current throw.
    \item Generates a new trajectory and continues execution.
\end{enumerate}

The total re-planning latency (single-target planning + multi-target coordination + trajectory generation) remains under 100~ms with Naive search, enabling smooth adaptation without stopping the robot. This reactive capability is only possible because of the computational efficiency of our framework---full Greedy search would require stopping for over a minute to re-plan.

\subsection{Discussion: Why Not Precompute Everything?}
A natural question is whether the entire multi-target coordination problem can be solved offline. For the two-object case, we discretize the target workspace (0.05\,m resolution in $xy$, 0.04\,m in $z$, annular region $1.25 \leq \|{(x,y)}\| \leq 1.4$\,m) and solve the full Greedy search for all valid target pairs, yielding $\sim$6.4M precomputed configuration pairs (6D key: two target positions; 28D value: joint angles and velocities for both throws). At runtime, a KD-tree nearest-neighbor query retrieves the closest precomputed solution. This achieves the fastest online planning ($<5$\,ms, dominated by trajectory generation), but has three limitations: (1)~it requires hours of offline computation; (2)~it introduces discretization error, as the retrieved solution corresponds to the nearest grid point rather than the exact query target; and (3)~it does not scale beyond two objects---for three objects, the precomputation space grows to $\sim$16 billion combinations, making brute-force offline solving infeasible. 

Table~\ref{tab:summary} summarizes the trade-offs across all strategies.  Note that the planning times reported there refer to multi-target coordination; single-target solution generation adds a fixed cost of under 5~ms per target, independent of the coordination strategy. Real-time operation is therefore a property of the strategy rather than the planner as a whole: single-target generation is always real-time, and only Naive or precomputed coordination preserves real-time end-to-end performance for reactive use, whereas the Greedy variants target offline or latency-tolerant planning where execution optimality is paramount.

\begin{table}[t]
    \centering
    \caption{Search strategy comparison}
    \label{tab:summary}
    \begin{tabular}{lcccl}
        \hline
        Method & Planning & Execution & Arbitrary & Scales to \\
               & time     & quality   & targets?  & $N$ objects? \\
        \hline
        Naive          & $<100$~ms  & Baseline  & Yes & Yes \\
        Greedy-$K$     & 0.1--20~s  & Improved  & Yes & Yes \\
        Greedy (full)  & $>60$~s    & Best      & Yes & Yes \\
        Precomputed    & $<5$~ms    & Best$^*$  & No  & No$^\dagger$ \\
        \hline
    \end{tabular}
    \vspace{1mm}
    
    {\footnotesize $^*$Execution quality matches full Greedy for nearby grid points but is subject to discretization error bounded by the grid resolution ($\sim$0.025\,m).\\
    $^\dagger$Currently implemented for 2 objects only; see text.}
\end{table}

\section{Discussion and Future Work}

\subsection{Synergy between grasping and throwing planning}
Scaling to three or more objects with the current hand proved impractical due to workspace limitations and inter-object interference during release.

We note that this grasping limitation is not fundamental to our planning framework. Recent advances in dexterous hand design~\cite{gao2026detachable} offer improved multi-object manipulation capabilities. Additionally, specialized gripper mechanisms with object storage in the palm~\cite{eom2024mogrip} provide an alternative approach where multiple objects can be stored and released sequentially. Our MOMT planning framework readily extends to such hardware, as it only requires knowledge of the object positions relative to the end-effector at release time. The planning of grasp configurations to minimize inter-object interference during release remains an interesting direction for future work.

\subsection{Collision-Free Motion Planning}
The current framework assumes an obstacle-free workspace. Extending it to cluttered environments requires integrating collision constraints into trajectory optimization while preserving the computational efficiency that enables reactive re-planning.
\subsection{Flight Dynamics}
We model objects as point masses in free flight (Assumption~2), which holds well for the compact, dense blocks used in our experiments but degrades for light or aerodynamically significant objects. Since the BRT is derived directly from the flight model, relaxing this assumption requires only recomputing the BRT by numerically integrating the true dynamics, including drag and spin~\cite{lubarda2022review}, while the online matching and multi-target coordination remain unchanged.

\section{Conclusions}
\label{sec:conclusions}
We presented a unified planning framework for multi-object multi-target (MOMT) dexterous throwing with fixed-base manipulators. Exploiting the rotational invariance of common manipulator kinematics, we reduced the feasibility problem to a compact 4D representation that supports online solution matching in under 5~ms per target, and introduced a pruned Greedy search that lets practitioners trade planning time against execution optimality.

On a Kuka iiwa 7 with an Allegro hand, the framework reduced execution time by 29\% over independent single-object planning and enabled reactive re-planning under dynamic target changes within 100~ms. To our knowledge, this is the first unified planning framework for MOMT throwing, validated in both simulation and real-world two-object experiments. Its modular decomposition into single-target planning and multi-target coordination provides a foundation for future capabilities such as dexterous grasp planning and collision-free trajectory optimization.



\bibliographystyle{IEEEtran}
\bibliography{references}

@article{liu2025transient,
  title={On Transient Release Dynamics in Robot Throwing: A Sliding Pivot Model},
  author={Liu, Yang and Billard, Aude},
  journal={IEEE Transactions on Robotics},
  volume={42},
  pages={132--151},
  year={2025},
  publisher={IEEE}
}

@inproceedings{liu2025learning,
  title={Learning to Throw-Flip},
  author={Liu, Yang and Da Costa, Bruno and Billard, Aude},
  booktitle={2025 IEEE/RSJ International Conference on Intelligent Robots and Systems (IROS)},
  pages={3209--3216},
  year={2025},
  organization={IEEE}
}

@article{li2024multigrasp,
  title={Grasp multiple objects with one hand},
  author={Li, Yuyang and Liu, Bo and Geng, Yiran and Li, Puhao and Yang, Yaodong and Zhu, Yixin and Liu, Tengyu and Huang, Siyuan},
  journal={IEEE Robotics and Automation Letters},
  volume={9},
  number={5},
  pages={4027--4034},
  year={2024},
  publisher={IEEE}
}

@inproceedings{oztop2006dexterous,
  title={Dexterous skills transfer by extending human body schema to a robotic hand},
  author={Oztop, Erhan and Lin, Li-heng and Kawato, Mitsuo and Cheng, Gordon},
  booktitle={2006 6th IEEE-RAS international conference on humanoid robots},
  pages={82--87},
  year={2006},
  organization={IEEE}
}

@article{gao2026detachable,
  title={A detachable crawling robotic hand},
  author={Gao, Xiao and Yao, Kunpeng and Junge, Kai and Hughes, Josie and Billard, Aude},
  journal={Nature Communications},
  volume={17},
  number={1},
  pages={428},
  year={2026},
  publisher={Nature Publishing Group UK London}
}

@article{eom2024mogrip,
  title={MOGrip: Gripper for multiobject grasping in pick-and-place tasks using translational movements of fingers},
  author={Eom, Jaemin and Yu, Sung Yol and Kim, Woongbae and Park, Chunghoon and Lee, Kristine Yoonseo and Cho, Kyu-Jin},
  journal={Science Robotics},
  volume={9},
  number={97},
  pages={eado3939},
  year={2024},
  publisher={American Association for the Advancement of Science}
}

@article{ma2025learning,
  title={Learning Accurate Whole-body Throwing with High-frequency Residual Policy and Pullback Tube Acceleration},
  author={Ma, Yuntao and Liu, Yang and Qu, Kaixian and Hutter, Marco},
  journal={arXiv preprint arXiv:2506.16986},
  year={2025}
}

@article{munn2024whole,
  title={Whole-Body Dynamic Throwing with Legged Manipulators},
  author={Munn, Humphrey and Tidd, Brendan and B{\"o}hm, Peter and Gallagher, Marcus and Howard, David},
  journal={arXiv preprint arXiv:2410.05681},
  year={2024}
}

@article{lubarda2022review,
  title={A review of the analysis of wind-influenced projectile motion in the presence of linear and nonlinear drag force},
  author={Lubarda, Marko V and Lubarda, Vlado A},
  journal={Archive of Applied Mechanics},
  volume={92},
  number={7},
  pages={1997--2017},
  year={2022},
  publisher={Springer}
}

@article{yao2023exploiting,
  title={Exploiting kinematic redundancy for robotic grasping of multiple objects},
  author={Yao, Kunpeng and Billard, Aude},
  journal={IEEE Transactions on Robotics},
  volume={39},
  number={3},
  pages={1982--2002},
  year={2023},
  publisher={IEEE}
}

@article{lynch1999dynamic,
  title={Dynamic nonprehensile manipulation: Controllability, planning, and experiments},
  author={Lynch, Kevin M and Mason, Matthew T},
  journal={The International Journal of Robotics Research},
  volume={18},
  number={1},
  pages={64--92},
  year={1999},
  publisher={SAGE Publications}
}

@article{zeng2020tossingbot,
  title={Tossingbot: Learning to throw arbitrary objects with residual physics},
  author={Zeng, Andy and Song, Shuran and Lee, Johnny and Rodriguez, Alberto and Funkhouser, Thomas},
  journal={IEEE Transactions on Robotics},
  volume={36},
  number={4},
  pages={1307--1319},
  year={2020},
  publisher={IEEE}
}

@article{zermane2024planning,
  title={Planning Impact-Driven Logistic Tasks},
  author={Zermane, Ahmed and Dehio, Niels and Kheddar, Abderrahmane},
  journal={IEEE Robotics and Automation Letters},
  year={2024},
  publisher={IEEE}
}

@inproceedings{werner2024dynamic,
  author       = {Lennart Werner and
                  Fang Nan and
                  Pol Eyschen and
                  Filippo A. Spinelli and
                  Hongyi Yang and
                  Marco Hutter},
  title        = {Dynamic Throwing with Robotic Material Handling Machines},
  booktitle    = {2024 {IEEE/RSJ} International Conference on Intelligent Robots and Systems},
  pages        = {98--104},
  publisher    = {{IEEE}},
  year         = {2024},
}

@inproceedings{kasaei2023throwing,
  title={Throwing objects into a moving basket while avoiding obstacles},
  author={Kasaei, Hamidreza and Kasaei, Mohammadreza},
  booktitle={2023 IEEE International Conference on Robotics and Automation (ICRA)},
  pages={3051--3057},
  year={2023},
  organization={IEEE}
}

@inproceedings{zhang2012sampling,
  title={Sampling-based motion planning with dynamic intermediate state objectives: Application to throwing},
  author={Zhang, Yajia and Luo, Jingru and Hauser, Kris},
  booktitle={2012 IEEE International Conference on Robotics and Automation},
  pages={2551--2556},
  year={2012},
  organization={IEEE}
}

@article{aboaf1987task,
  title={Task-level robot learning: Ball throwing},
  author={Aboaf, Eric W and Atkeson, Christopher G and Reinkensmeyer, David J},
  year={1987}
}

@inproceedings{pekarovskiy2013optimal,
  title={Optimal control goal manifolds for planar nonprehensile throwing},
  author={Pekarovskiy, Alexander and Buss, Martin},
  booktitle={2013 IEEE/RSJ International Conference on Intelligent Robots and Systems},
  pages={4518--4524},
  year={2013},
  organization={IEEE}
}

@inproceedings{liu2022solution,
  title={A solution to adaptive mobile manipulator throwing},
  author={Liu, Yang and Nayak, Aradhana and Billard, Aude},
  booktitle={2022 Ieee/Rsj International Conference On Intelligent Robots And Systems (Iros)},
  pages={1625--1632},
  year={2022},
  organization={IEEE}
}

@article{liu2024tube,
  title={Tube acceleration: robust dexterous throwing against release uncertainty},
  author={Liu, Yang and Billard, Aude},
  journal={IEEE Transactions on Robotics},
  year={2024},
  publisher={IEEE}
}

@inproceedings{lombai2009throwing,
  title={Throwing motion generation using nonlinear optimization on a 6-degree-of-freedom robot manipulator},
  author={Lombai, Ferenc and Szederk{\'e}nyi, G{\'a}bor},
  booktitle={2009 IEEE International Conference on Mechatronics},
  pages={1--6},
  year={2009},
  organization={IEEE}
}

@article{berscheid2021jerk,
  title={Jerk-limited Real-time Trajectory Generation with Arbitrary Target States},
  author={Berscheid, Lars and Kr{\"o}ger, Torsten},
  journal={Robotics: Science and Systems XVII},
  year={2021}
}

@inproceedings{senoo2008high,
  title={High-speed throwing motion based on kinetic chain approach},
  author={Senoo, Taku and Namiki, Akio and Ishikawa, Masatoshi},
  booktitle={2008 IEEE/RSJ International Conference on Intelligent Robots and Systems},
  pages={3206--3211},
  year={2008},
  organization={IEEE}
}

\end{document}